\documentclass[runningheads,a4paper]{llncs}

\usepackage[margin=2.5cm]{geometry}

\usepackage{amssymb}
\usepackage{graphicx}
\usepackage{microtype}
\usepackage{amsmath}
\usepackage{wrapfig}
\usepackage{subfig}
\usepackage{float}

\usepackage{algorithm}
\usepackage[noend]{algpseudocode}

\usepackage{url}
\urldef{\mailsa}\path|christian.perone@gmail.com|

\begin{document}

\setlength\intextsep{0pt}

\mainmatter  

\title{Towards ECDSA key derivation from deep embeddings for novel Blockchain applications}

\titlerunning{Towards ECDSA key derivation from deep embeddings for novel Blockchain applications}

%
%
\author{Christian S. Perone}

\authorrunning{Towards ECDSA key derivation from deep embeddings for novel Blockchain applications}

\institute{
\mailsa\\
\url{http://blog.christianperone.com}}

%
%

\maketitle

\begin{abstract}
In this work, we propose a straightforward method to derive Elliptic Curve Digital Signature Algorithm (ECDSA) key pairs from embeddings created using Deep Learning and Metric Learning approaches. We also show that these keys allows the derivation of cryptocurrencies (such as Bitcoin) addresses that can be used to transfer and receive funds, allowing novel Blockchain-based applications that can be used to transfer funds or data directly to domains such as image, text, sound or any other domain where Deep Learning can extract high-quality embeddings; providing thus a novel integration between the properties of the Blockchain-based technologies such as trust minimization and decentralization together with the high-quality learned representations from Deep Learning techniques. 

\end{abstract}

%



\section{Introduction}
Decentralized cryptocurrencies such as Bitcoin \cite{Nakamoto2008}, enabled a wide spectrum of novel applications. Based on the Blockchain, the underlying distributed core of Bitcoin, applications ranging from smart contracts, verifiable data, and transactions, audit systems, among others, made a significant impact on the way that we can make transactions and build systems with trust minimization or no third-party involved at all. Supported and advocated by a vibrant community, Blockchain is a technology that can be applied to many contexts beyond cryptocurrency and financial assets \cite{Swan2015}.

In the recent years, we also saw the fast development of Deep Learning \cite{LeCun2015a}, especially in the Computer Vision area, where we were able to see major milestones such as the well known AlexNet \cite{Krizhevsky2012a}, VGGNet \cite{Simonyan2014} GoogleNet \cite{Szegedy2014} and more recently the ResNets \cite{He}.
The representation of high-dimensional data, such as natural images, by low-dimensional embeddings learned using deep neural networks with multiple hierarchical features, allowed not only image classification, but also many different applications such as visual search \cite{Jing2015} \cite{Bell2015}, image hashing \cite{Lin2015} \cite{Liong2015} \cite{Cao2017}, image captioning \cite{Vinyals2016} and many other applications.

Although the synergy between the Blockchain or Blockchain-based technology and Machine Learning/AI was mentioned in the health context, such as in \cite{Swan2015} for the application of secure large-scale data management mechanism to coordinate the information of individuals, and more recently in \cite{Suleyman2017a} for the application of a verifiable data audit system, there are very few published works or implemented applications exploring this integration between learned representations and the Blockchain beyond the trend of Machine Learning model auditing. In \cite{Swan2015}, the author also speculatively cites an application of the Blockchain and AI systems, but for the context of communication between AI entities in order to conduct certain transactions.

In this context, our work focuses on the exploration of this potential synergy between representation learning and the Blockchain by means of novel practical application example and experimentation.

Our contributions are as follows: first, we focus on the usage of Convolutional Neural Networks (CNNs) \cite{LeCun2004} and metric learning \cite{Chopra2005} to produce embeddings for the purpose of cryptography key-pair derivation. Later, we develop a method for deriving Elliptic Curve Digital Signature Algorithm (ECDSA) key-pairs from deep embeddings extracted from natural images, as well as an application example using the Blockchain technology that will allow us to create transactions that can only be redeemed by individuals owning the same similar images or the transaction owner. The ECDSA is used by the Bitcoin to promote cryptography guarantees for the transactions, ensuring that funds can only be spent by their owners, thus we use the same mechanisms in this work to promote cryptographic guarantees for the fund transactions. Later, we discuss future extensions of this work and other potential applications of this integration between representation learning and cryptocurrencies/Blockchain.

\newpage

\section{Related Work}
As we mentioned in the previous section, there are very few published works related to AI/Machine Learning and Blockchain/cryptocurrencies.
In \cite{Swan2015}, the author explores some Blockchain applications in a highly-speculative manner. One of the applications that the author cite is the usage of the Blockchain by AI entities as a mean to provide a permanent transparent public record that can be reviewed and inspected. 

In \cite{Swan2015a}, they cite an application called \emph{Monegraph}, created to provide property ownership of assets such as online graphics or digital media, where a user can pay a small network fee via namecoin and have its Twitter account and URL being placed on the Blockchain. However, they do not account for similarity issues. In \cite{Swan2015a}, they also cite a digital art and copyright protection project called \emph{Ascribe}, however, they use Machine Learning for similarity search while crawling the web, so there is no evidence of a link between digital art extracted features embedded on the Blockchain. Only the hash of the digital content and metadata is added to Blockchain.

Some works such as Crypto-Nets \cite{Xie2014} used homomorphic encryption to provide privacy-preserving guarantees, but no keys were derived from the embeddings since the goal was to provide privacy.

A thorough search of the relevant literature yielded no relevant similar work. To the best of our knowledge, this is the first study exploring the use of embeddings in the context of the Blockchain.

\section{Background}

\subsection{ECDSA curves and the secp256k1}
The Elliptic Curve Digital Signature Algorithm, also called ECDSA, is the technique used by the Bitcoin cryptocurrency to guarantee that the funds can only be used by their rightful owners. It is beyond the scope of this work to describe how ECDSA works, so we'll just describe the important concepts that will be mentioned in this work.
In order to use Elliptic Curve Cryptography (ECC), all parties must agree on all the elements defining the elliptic curve, the domain parameters. Many standard bodies published domain parameters of elliptic curves for different field sizes. The standard used by Bitcoin is called \emph{secp256k1} and was proposed by \cite{DanielR.L.Brown2010}, where it was constructed in a special way to allow efficient computation. The secp256k1's parameters were also selected in a predictable way, which significantly reduces the possibility that the curve's creator inserted any sort of back-door into the curve \cite{bitsecwiki}. 

\subsection{Bitcoin Addresses and Transactions}
\begin{wrapfigure}{r}{6.0cm}
	\centering
	\includegraphics[clip, trim=10cm 2cm 10cm 2cm,width=\textwidth,height=9cm,keepaspectratio]{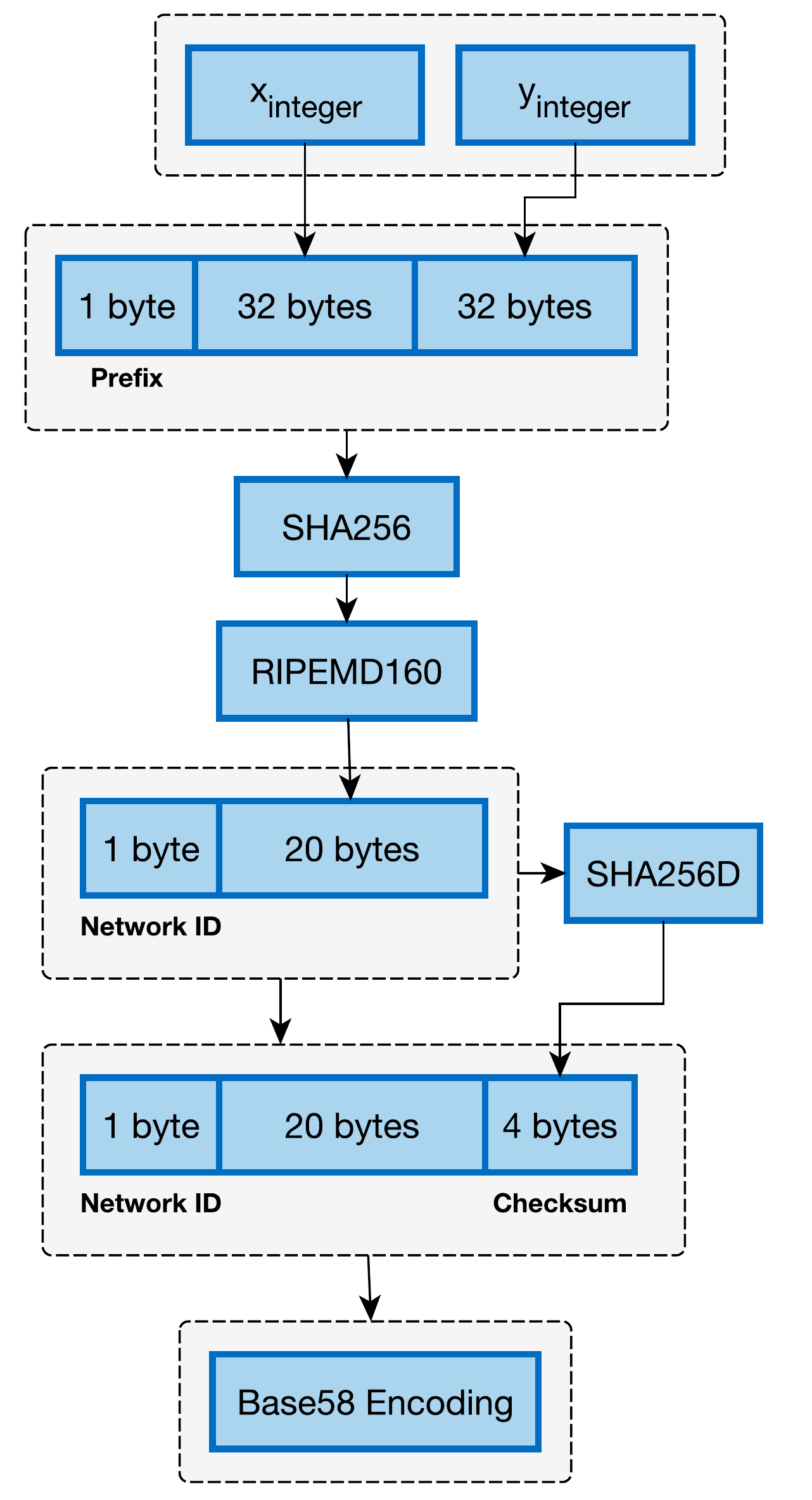}
	\caption{An overview of the Bitcoin address generation.}
	\label{fig:bitcoinaddress}
\end{wrapfigure}

A pseudo-random number generator (PRNG) is typically used in Bitcoin to generate the private key for the secp256k1, where a 256-bit number is randomly generated and then used as the secret exponent of the secp256k1 curve. It is important to note that not all 256-bit numbers are valid ECDSA private keys and this range of valid private keys is governed by the secp256k1 ECDSA standard, however, the upper bound of the valid exponent is very close to the largest 256-bit number. The main goal of this work is to derive this private key using deterministic representations created by Metric Learning over features extracted from deep Convolutional Neural Networks (CNNs) instead of using a pseudo-random number generator.

\subsubsection{Key pair and address generation.}

The process to create a Bitcoin address starts with the generation of the non-zero random integer $d$ private key and then a public key $Q = dG$ is derived, where $G$ is the base point parameter. After that, the process to generate a Bitcoin address can be seen in the Figure~\ref{fig:bitcoinaddress}, where basically, there are two rounds of two different hash functions (SHA-256 and RIPEMD-160) over the public key, after that the SHA256D is used to create a checksum and then the address is encoded into a Base58 encoding scheme. What is important to note here, is that the final Bitcoin address isn't the public key itself, but the hash of the public key.
	
\subsubsection{Transactions.} There are two main standard transaction types that are used on the Bitcoin network nowadays. The first one is called \emph{``Pay To Public Key Hash''} or simply P2PKH. This transaction, as the name states, pays for the owner of the public key that hashes to the specified address. Since the sender can't provide the public key, when redeeming the funds of the transaction, the owner of the public key must provide his public key that hashes to the specified address and also the signature.

The other commonly used transaction type is called \emph{``Pay To Script Hash''}, where the responsibility for supplying the conditions to redeem a transaction is moved from the sender of the funds to the redeemer of the funds. This transaction type allows flexibility for specifying larger and complex scripts without letting the sender to worry about it or even know the script contents.

\section{Methods and Materials}
Learning a good representation that can be used to represent different natural images is an important task. In our context, we not only want to learn a good representation with similarity properties but also a representation that can be \emph{binarized} and used as the deterministic seed for the ECDSA private key derivation.

There has been considerable research on distance metric learning over the past few years \cite{Yang2006} \cite{Bellet2015}. The goal of metric learning is to adapt some pairwise real-valued metric function, using the information from training examples \cite{Bellet2015}. Most methods learn the metric in a weakly-supervised way from pairs or triplets \cite{Schroff2015}. Since the creation of an architecture to learn the embeddings isn't the main goal of this work, we used a simple siamese network \cite{Chopra2005} with contrastive loss to learn the representations such that the squared L2 distances in the embedding space directly correspond to image similarity.

Since the indefinite contraction of the pairs was found to be detrimental to the training dynamics \cite{Lin2015}, we employed the same proposed double-margin loss as seen in \cite{Lin2015}:

\begin{equation}
\label{eq:siamese2}
\begin{split}
\mathcal{L}(\mathbf{z}_\alpha,\mathbf{z}_\beta) = &y \max(\lVert \mathbf{z}_\alpha - \mathbf{z}_\beta \rVert_2^2 - m_1, 0) + (1-y) \max(m_2 - \lVert \mathbf{z}_\alpha - \mathbf{z}_\beta \rVert_2^2, 0)
\end{split}
\end{equation}

Where the $m_1$ and $m_2$ are two different margins that we choose experimentally, $\mathbf{z}_\alpha,\mathbf{z}_\beta$ are the training pairs and $y$ is equal to 1 when the pairs are a genuine pair or 0 when they are impostors.

As we mentioned before, we need a representation that when \emph{binarized} will have an approximate performance result of the original representation for retrieval. To accomplish such representation property, we applied a triangular regularization prior on the activations of the last layer of the the network (the layer we use to extract embeddings) described in the element-wise equation below:
	
\begin{equation} 
  f_{reg}(\psi) = \lambda * (1 - | (\sigma(x) * 2) - 1|)
\end{equation}


\begin{wrapfigure}{r}{6.0cm}
	\centering
	\includegraphics[width=6cm]{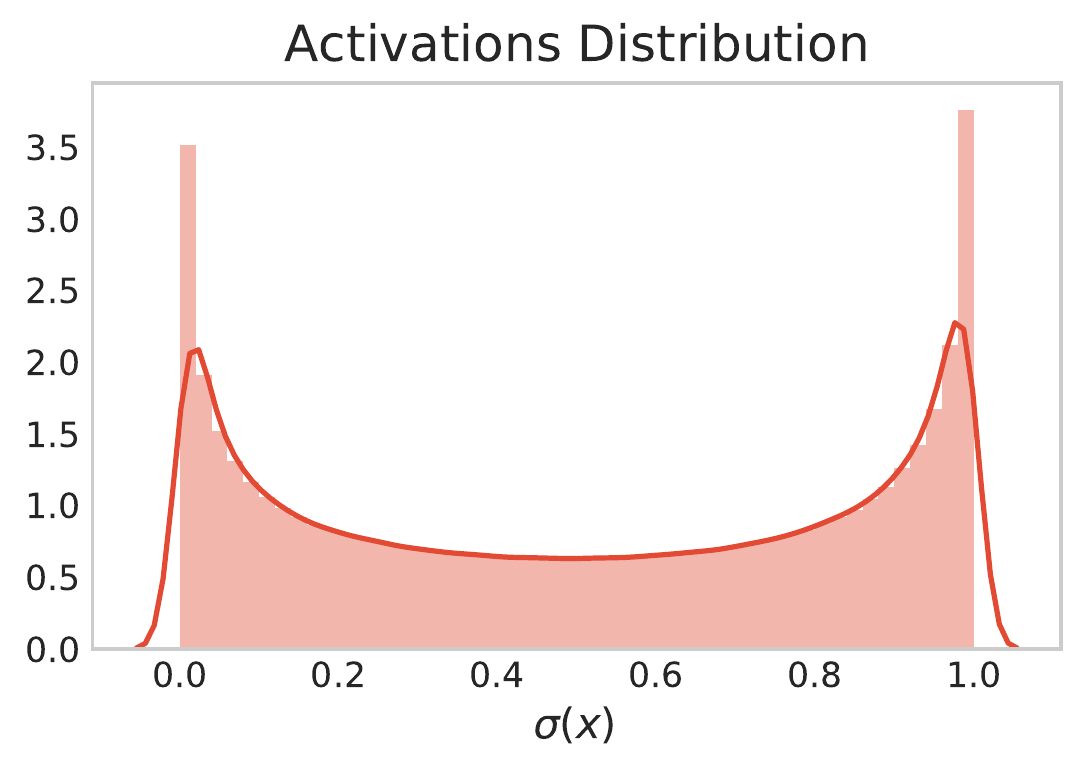}
	\caption{Activation distribution after training using the proposed regularization.}
	\label{fig:act-distr}
\end{wrapfigure}

Where $\sigma(\cdot)$ is the sigmoid activation function, $x$ is the pre-activation and $\lambda$ is a balancing term. We used a pre-trained ResNet-50 \cite{He2016} as the network architecture for feature extraction and we added a fully-connected layer of 256 units as the last layer. The network training process was done using random positive and negative sample pairs from the ImageNet dataset. After training the network with the proposed regularization (eq. 1), we can see that the activations lie mostly on the upper and lower bounds of the sigmoid activation as seen in Figure \ref{fig:act-distr}.

\begin{figure}
	\centering
	\includegraphics[width=\textwidth]{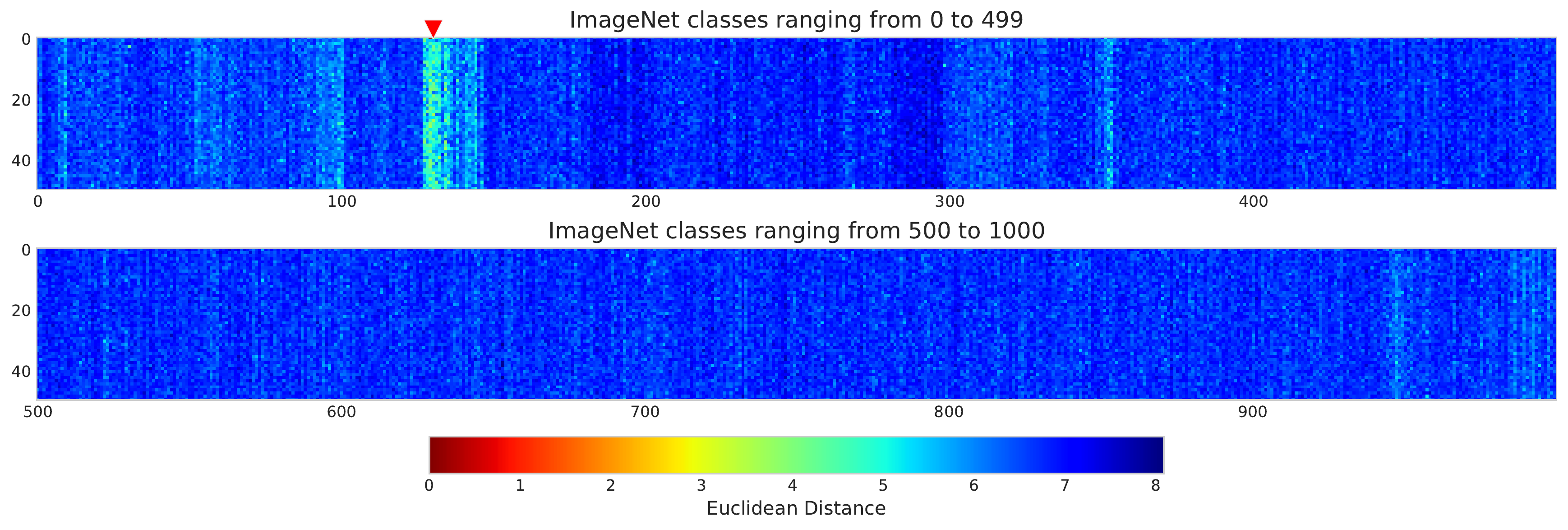}
	\caption{Euclidean distance similarity for one sample.}
	\label{fig:matrixdistance}
\end{figure}

In the Figure \ref{fig:matrixdistance}, we show the Euclidean distance between the feature vector of a single random sample from the ImageNet belonging to the class 129, which represents the Spoonbill bird. As we can see, the most similar classes are also close to the Spoonbill class, that also represent birds such as Flamingos, hence the lower Euclidean distance.

\begin{figure}
	\centering
	\subfloat[tSNE of non-binarized features.]{{\includegraphics[width=7.5cm]{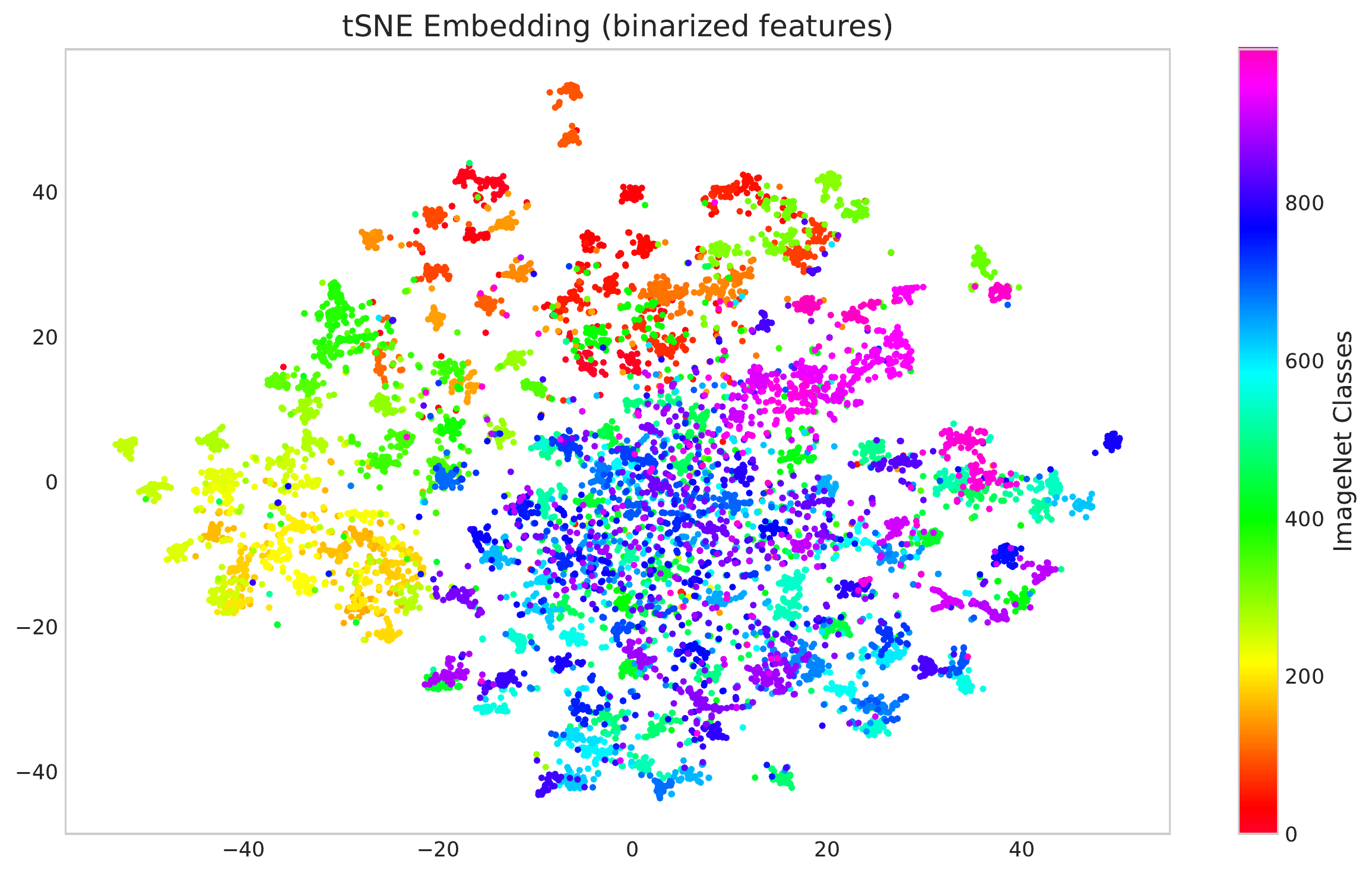} }}%
	\qquad
	\subfloat[tSNE of binarized features.]{{\includegraphics[width=7.5cm]{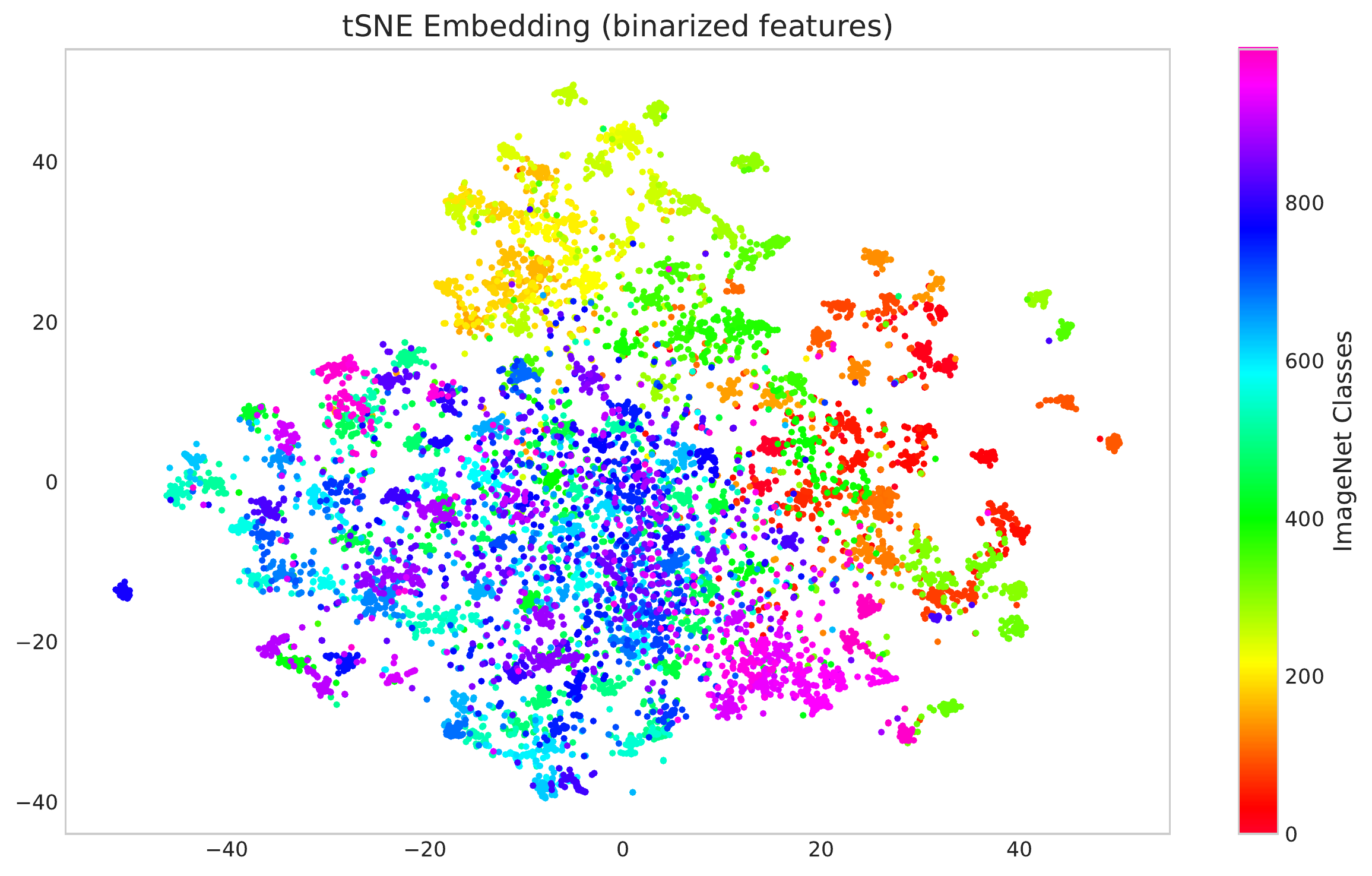} }}%
	\caption{tSNE embeddings.}%
	\label{fig:tsnebinarized}%
\end{figure}

In Figure \ref{fig:tsnebinarized}, we show the t-SNE\cite{VanDerMaaten2008} visualization of the feature vectors before and after the binarization, where we can see that similar clustering patterns are preserved after the binarization process.

\begin{figure}  
	\centering
	\includegraphics[clip, trim=1cm 8.5cm 0cm 4cm,width=\textwidth]{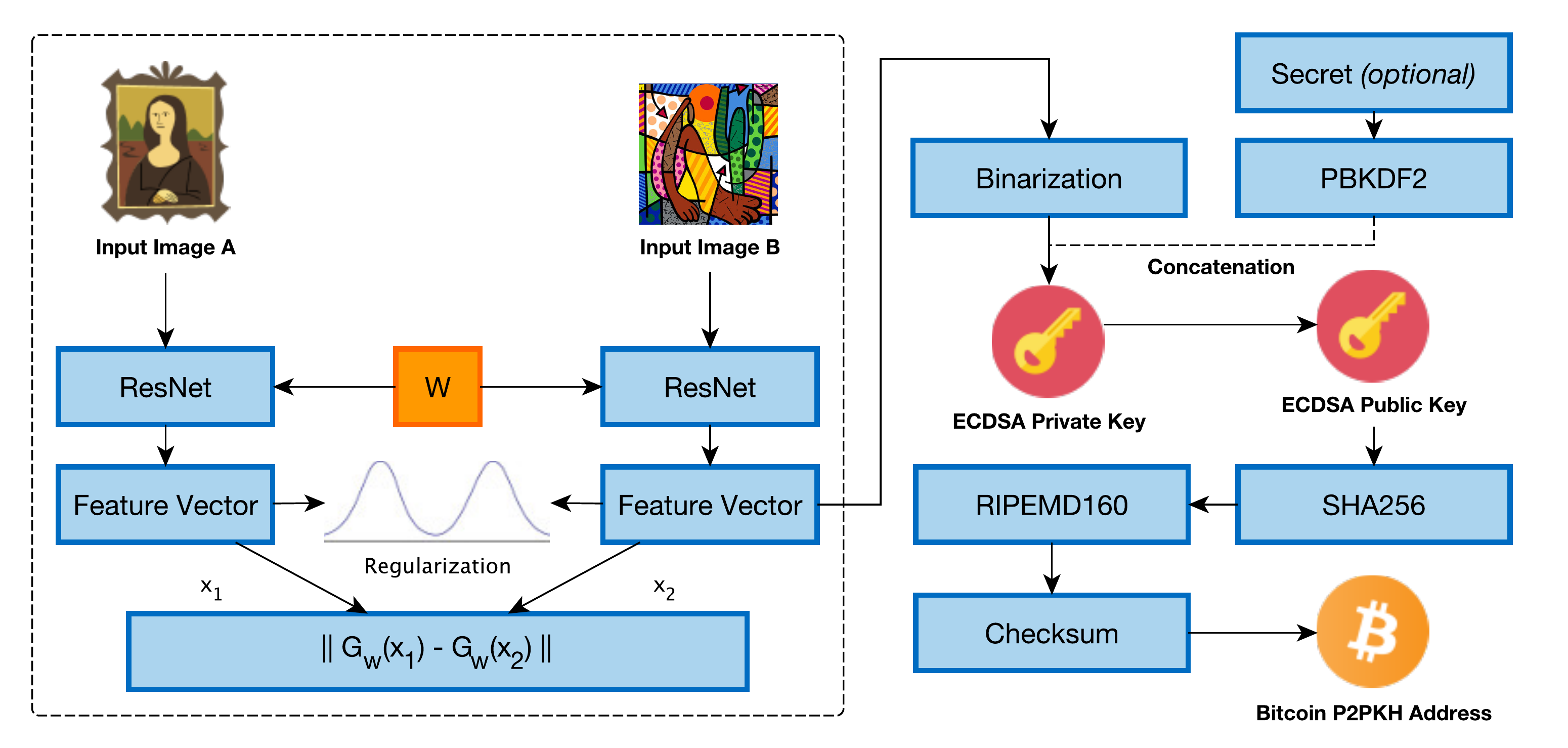}
	\caption{An overview of the training procedure and method to derive the ECDSA key pair using the representations learned by metric learning using a CNN.}
	\label{fig:flow}
\end{figure}

After training the network, we use it to extract feature representations from images and then binarize the feature vector $\vec{v} \in \mathbb{R}^{256}$  into a binary vector $\vec{v} \in \mathbb{Z}^{256}_{2}$, where $\mathbb{Z}_{2} = [0,1]$. After that, we use the binary feature vector as the secret for deriving the public key using the secp256k1 base point parameters and then we follow the same Bitcoin address generation protocol where a series of hashed are applied over the public key coordinates to derive the final Bitcoin address. 
In Figure \ref{fig:flow} we show an overview of the complete described flow.


The algorithm for deriving the public key can be seen in the Algorithm 1, where we describe the steps to derive the public key from the feature vectors.
\\

\begin{algorithm}[H]
	
	\caption{ECDSA key-pair derivation algorithm}\label{euclid}
	\begin{algorithmic}[1]
		\Procedure{DeriveKeyPairs}{}
		\State $\textit{network} \gets \text{pre-trained siamese network}$
		\State $\textit{image} \gets \text{input image}$
		\State $\textit{repr} \gets network(image)$ \Comment{Forward pass to get representation $\textit{repr} \in \mathbb{R}^{128}$}
		\State $\textit{secret} \gets binarize(repr)$ \Comment{Binarize features, $repr\_bin \in \mathbb{Z}^{256}_{2}$, where $\mathbb{Z}_{2} = [0,1]$}
		\State $\textit{opt\_passwd} \gets PBKDF2(passwd)$ \Comment{Use a password (optional)}
		\State $\textit{priv\_key} \gets \textit{secret}  \, \| \, \textit{opt\_passwd}$ \Comment{Concatenate the secret with the password (optional)}
		
		\State $\textit{pub\_key} \gets \textit{priv\_key}  \otimes G$ \Comment{Derive the ECDSA public key using secp256k1}	
		\EndProcedure
	\end{algorithmic}
\end{algorithm}

As we can see from the Algorithm 1, the procedure to derive the key is straightforward. First we feed the desired image into the network to compute the forward pass, after that we use the extracted features and binarize them to create the ECDSA secret. As optional stage, we can concatenate (taking bit-length into consideration) the derived secret with a user password where a PBKDF2 function is applied to avoid some attack vectors such as rainbow tables, and then finally we use the ECDSA private-key to derive the public-key by doing the field multiplication with the secp256k1 parameters.

After creating the public-key, we can derive a Bitcoin address using the scheme shown in Figure \ref{fig:bitcoinaddress}. With the Bitcoin address in hands, we can transfer funds to this addres in a way that it can be redeemed only when a similar image on the learned low-dimensional manifold instead of other pixel space or traditional image hashing is used to derive the private-key, or if using the optional password: only someone with the password and similar image will be able to redeem the transfered funds.

\section{Discussion}

As we saw, through a very simple process, we can create an ECDSA key-pair that is derived not from a random seed but from a binarized feature vector representing an image. This key-pair can be used to sign or encrypt data and in our context, can be used to derive Bitcoin addresses that only owners of the same or \textit{similar} images (on the learned low-dimensional manifold) can redeem the contents of the transaction. By using a P2PKH (Pay to Public Key Hash) transaction, one can transfer cryptocurrency funds to the image.

The described technique is not limited to the Bitcoin cryptocurrency but can be used for any Blockchain-based technology that relies on the ECDSA. Although we provided a very simple siamese architecture with a simple regularization mechanism to keep the Euclidean distance on the low dimensional manifold, our technique isn't limited to this particular architecture, therefore other "deep hash" techniques such as HashNets \cite{Cao2017} can be employed to create the representations that will be later used to derive the ECDSA key-pair. Our method is also not limited to the image domain, since the representations can be created from text or even other different input domains. In the Bitcoin context, users can transfer funds to natural images, texts, sounds or any other domain where Deep Learning can successfully extract high-quality embeddings, providing a new door for many potential applications based on the Blockchain technology.

To conclude, we showed in this work an approach to derive ECDSA keys from feature vectors, allowing many novel applications where users can transfer funds or store decentralized data on natural images, text and sounds. To the best of our knowledge, this is the first study to propose a concrete method to allow the integration between Deep Learning and the Blockchain technology. Further studies are certainly required in order to assess the potential attack vectors to this approach, however we believe that this integration would certainly open the door to novel applications that can provide trust minimization for transfering data or funds to many different data domains.

\section{Additional Information}
\textbf{Software}: To develop this work, we used the Keras framework using the TensorFlow backend. To create the ECDSA keys and derive Bitcoin addresses we used the Protocoin\footnote{http://protocoin.readthedocs.io/} framework for creating the ECDSA keys and Bitcoin addresses.

\noindent \textbf{Disclaimer}: This method is provided by the author and contributors "as is" and any express or implied warranties, including, but not limited to, the implied warranties of merchantability and fitness for a particular purpose are disclaimed. In no event shall the copyright owner or contributors be liable for any direct, indirect, incidental, special, exemplary, or consequential damages (including, but not limited to, procurement of substitute goods or services; loss of use, data, or profits; or business interruption) however caused and on any theory of liability, whether in contract, strict liability, or tort (including negligence or otherwise) arising in any way out of the use of this method, even if advised of the possibility of such damage. The author isn't responsible for any loss of cryptocurrencies or any financial asset due to the use of the method described in this article.


\bibliography{../gmseg/library}

\begin{thebibliography}{10}
\providecommand{\url}[1]{\texttt{#1}}
\providecommand{\urlprefix}{URL }

\bibitem{bitsecwiki}
{Secp256k1 - Bitcoin Wiki}, \url{https://en.bitcoin.it/wiki/Secp256k1}

\bibitem{Bell2015}
Bell, S.: {Learning visual similarity for product design with convolutional
  neural networks}. Siggraph  34(4),  1--9 (2015)

\bibitem{Bellet2015}
Bellet, A., Habrard, A., Sebban, M.: {A Survey on Metric Learning for Feature
  Vectors and Structured Data}. Bmvc2015 p.~57 (2015)

\bibitem{Cao2017}
Cao, Z., Long, M., Wang, J., Yu, P.S.: {HashNet: Deep Learning to Hash by
  Continuation}. arXiv  (2017)

\bibitem{Chopra2005}
Chopra, S., Hadsell, R., Y., L.: {Learning a similiarty metric
  discriminatively, with application to face verification}. Proceedings of IEEE
  Conference on Computer Vision and Pattern Recognition pp. 349--356 (2005)

\bibitem{DanielR.L.Brown2010}
{Daniel R. L. Brown}: {Standards for Efficient Cryptography 2 (SEC 2) :
  Recommended Elliptic Curve Domain Parameters}. Standards for Efficient
  Cryptography p.~37 (2010)

\bibitem{He2016}
He, K., Zhang, X., Ren, S., Sun, J.: {Deep Residual Learning for Image
  Recognition}. In: 2016 IEEE Conference on Computer Vision and Pattern
  Recognition (CVPR). pp. 770--778 (2016)

\bibitem{He}
He, K., Zhang, X., Ren, S., Sun, J.: {Identity mappings in deep residual
  networks}. Lecture Notes in Computer Science (including subseries Lecture
  Notes in Artificial Intelligence and Lecture Notes in Bioinformatics)  9908
  LNCS,  630--645 (2016)

\bibitem{Jing2015}
Jing, Y., Liu, D., Kislyuk, D., Zhai, A., Xu, J., Donahue, J., Tavel, S.:
  {Visual Search at Pinterest}. Proceedings of the 21th ACM SIGKDD
  International Conference on Knowledge Discovery and Data Mining pp.
  1889--1898 (2015)

\bibitem{Krizhevsky2012a}
Krizhevsky, A., Sutskever, I., Hinton, G.E.: {ImageNet Classification with Deep
  Convolutional Neural Networks}. Advances In Neural Information Processing
  Systems pp. 1--9 (2012)

\bibitem{LeCun2015a}
LeCun, Y., Bengio, Y., Hinton, G., Y., L., Y., B., G., H.: {Deep learning}.
  Nature  521(7553),  436--444 (2015)

\bibitem{LeCun2004}
LeCun, Y., Huang, F.J.H.F.J., Bottou, L.: {Learning Methods for Generic Object
  Recognition with Invariance to Pose and Lighting}. Computer Vision and
  Pattern Recognition, 2004. CVPR 2004. Proceedings of the 2004 IEEE Computer
  Society Conference on  2,  II--97 -- 104 (2004)

\bibitem{Lin2015}
Lin, J., Morere, O., Chandrasekhar, V.: {DeepHash: Getting Regularization,
  Depth and Fine-Tuning Right}. arXiv preprint arXiv: {\ldots} p.~20 (2015)

\bibitem{Liong2015}
Liong, V.E., Lu, J., Wang, G., Moulin, P., Zhou, J.: {Deep hashing for compact
  binary codes learning}. In: Proceedings of the IEEE Computer Society
  Conference on Computer Vision and Pattern Recognition. vol. 07-12-June, pp.
  2475--2483 (2015)

\bibitem{Nakamoto2008}
Nakamoto, S.: {Bitcoin: A Peer-to-Peer Electronic Cash System}. Www.Bitcoin.Org
  p.~9 (2008), \url{https://bitcoin.org/bitcoin.pdf}

\bibitem{Schroff2015}
Schroff, F., Kalenichenko, D., Philbin, J.: {FaceNet: A unified embedding for
  face recognition and clustering}. Proceedings of the IEEE Computer Society
  Conference on Computer Vision and Pattern Recognition  07-12-June,  815--823
  (2015)

\bibitem{Simonyan2014}
Simonyan, K., Zisserman, A.: {Very Deep Convolutional Networks for Large-Scale
  Image Recognition}. ImageNet Challenge pp. 1--10 (2014)

\bibitem{Suleyman2017a}
Suleyman, M., Laurie, B.: {Trust, confidence and Verifiable Data Audit} (2017),
  \url{https://deepmind.com/blog/trust-confidence-verifiable-data-audit/}

\bibitem{Swan2015a}
Swan, M.: {Blockchain: blueprint for a new economy}. O'Reilly, first edit edn.
  (2015)

\bibitem{Swan2015}
Swan, M.: {Blockchain Thinking : the Brain as a Decentralized Autonomous
  Corporation [Commentary]} (2015)

\bibitem{Szegedy2014}
Szegedy, C., Liu, W., Jia, Y., Sermanet, P., Reed, S., Anguelov, D., Erhan, D.,
  Vanhoucke, V., Rabinovich, A., Hill, C., Arbor, A.: {Going Deeper with
  Convolutions}  (2014)

\bibitem{VanDerMaaten2008}
{Van Der Maaten}, L., Hinton, G.: {Visualizing Data using t-SNE}. Journal of
  Machine Learning Research  9,  2579--2605 (2008)

\bibitem{Vinyals2016}
Vinyals, O., Toshev, A., Bengio, S., Erhan, D.: {Show and Tell: Lessons learned
  from the 2015 MSCOCO Image Captioning Challenge}. TPAMI  99(PP),  1--1 (2016)

\bibitem{Xie2014}
Xie, P., Bilenko, M., Finley, T., Gilad-Bachrach, R., Lauter, K., Naehrig, M.:
  {Crypto-Nets: Neural Networks over Encrypted Data}  (2014),
  \url{https://arxiv.org/pdf/1412.6181.pdf http://arxiv.org/abs/1412.6181}

\bibitem{Yang2006}
Yang, L., Jin, R.: {Distance metric learning: A comprehensive survey}. Michigan
  State Universiy pp. 1--51 (2006)

\end{thebibliography}
\bibliographystyle{splncs03}


\end{document}